\documentclass[twocolumn]{article}

\usepackage{microtype}
\usepackage{graphicx}
\usepackage{subfigure}
\usepackage{booktabs} 
\usepackage{array}
\usepackage{multirow}
\usepackage{multicol}
\usepackage{hhline}
\usepackage{makecell}
\usepackage{libertine}

\usepackage{xcolor}
\usepackage{color}
\usepackage{colortbl}
\definecolor{black}{rgb}{0,0,0}
\definecolor{white}{rgb}{1,1,1}
\definecolor{darkred}{rgb}{0.5,0,0}
\definecolor{darkgreen}{rgb}{0,0.5,0}
\definecolor{darkblue}{rgb}{0,0,0.5}

\usepackage{amsmath}
\usepackage{amssymb}

\usepackage[a4paper, left=1in, right=1in, top=1in, bottom=1in]{geometry}

\usepackage{hyperref}

\title{Points2Vec: Unsupervised Object-level Feature Learning from Point Clouds}
\date{}
\usepackage{authblk}
\begin{document}

\author[1, 2]{Jo\"el Bachmann \thanks{Correspondence to joel@mantistechnologies.ch}}
\author[2]{Kenneth Blomqvist}
\author[2]{Julian F\"orster}
\author[2]{Roland Siegwart}

\affil[1]{Mantis Technologies}
\affil[2]{Autonomous Systems Lab, Swiss Federal Institute of Technology, Zurich, Switzerland}

\maketitle

\begin{abstract}
Unsupervised representation learning techniques, such as learning word embeddings, have had a significant impact on the field of natural language processing. Similar representation learning techniques have not yet become commonplace in the context of 3D vision. This, despite the fact that the physical 3D spaces have a similar semantic structure to bodies of text: words are surrounded by words that are semantically related, just like objects are surrounded by other objects that are similar in concept and usage.

In this work, we exploit this structure in learning semantically meaningful low dimensional vector representations of objects. We learn these vector representations by mining a dataset of scanned 3D spaces using an unsupervised algorithm. We represent objects as point clouds, a flexible and general representation for 3D data, which we encode into a vector representation. We show that using our method to include context increases the ability of a clustering algorithm to distinguish different semantic classes from each other. Furthermore, we show that our algorithm produces continuous and meaningful object embeddings through interpolation experiments. 
\end{abstract}

\section{Introduction}
\label{sec:introduction}
Unsupervised learning has seen much success in natural language processing \cite{mikolov13_WordRep, pennington2014glove, brown2020language}. Many current methods rely on models that have been pretrained on standard tasks. Others use representations learned ahead of time, such as word vectors. Largely, this is driven by massive amounts of text data that is available through the internet. Similarly in computer vision, pretrained models are used to improve generalization and to learn from a small amount of data. Methods have been developed to learn representations from unlabeled data, to speed up learning on downstream tasks \cite{jing19_survey}. This has been much less explored in the context of geometric 3D data.

Methods like word2vec \cite{mikolov13_WordRep} leverage the context in which a word appears to learn a semantic representation for each word. Not only can the meaning of a word be altered by its context, but words appearing as neighbors are also correlated. Words from the same semantic realm often appear close to each other in text or have similar neighbors. Just like words in bodies of text, objects in physical spaces are defined by the context in which they appear and are used. A hammer will likely be found near a screwdriver. A chair could be swapped with stool, without it appearing out of place. We build on this realization by designing an unsupervised approach to learning low dimensional object representations.

\begin{figure}
	\centering
	\includegraphics[width = 1\linewidth]{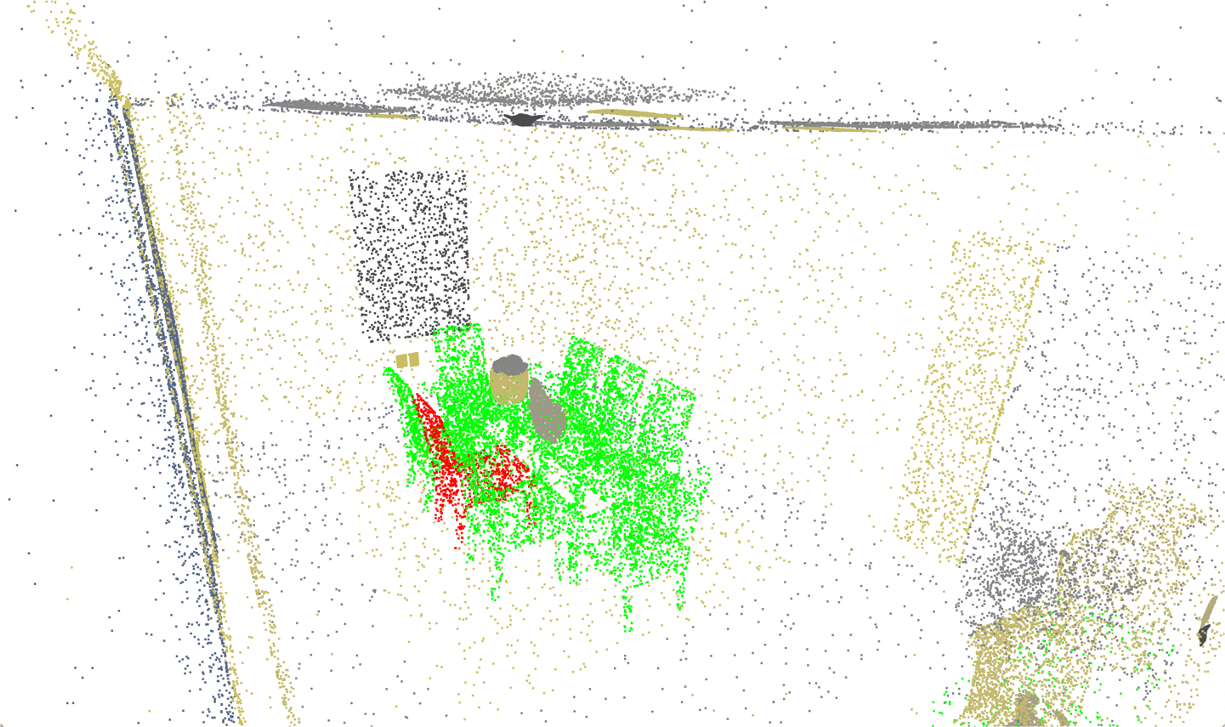}
	\caption{An application of our work is finding semantically similar objects in a scene. Here, we give our algorithm an instance (red), and ask it to search for the instances that are most similar to that instance (green). The algorithm correctly outputs all other chairs and the table.}
	\label{fig:use_case}
\end{figure}

Our goal is to find a mapping from 3D objects into a low-dimensional vector representation that is representative of the semantic meaning of the object – an embedding function. We base our approach on the idea that semantically relevant features of 3D objects are both dependent on the context in which they appear, as well as the geometry of the object. Within the low-dimensional representation, a chair in a kitchen, for instance, should be separable from an office chair, or conversely, pillows should occupy a nearby region in the semantic space to the region occupied by beds.

We choose to use point clouds as the representation for our objects and 3D geometry. The reason is threefold. Any 3D representation, such as a triangle mesh, a CAD model or a voxel grid, can be converted to a point cloud by sampling points on the surface of the encoded geometry. Point clouds can be obtained by increasingly common depth sensors, such as lidar scanners. Lastly, effective differentiable functions that operate on point clouds, have been developed, PointNet \cite{qi17_PointNet} being one such example.

In this paper we:
\begin{itemize}
	\item Propose an algorithm that embeds point clouds into a low dimensional vector space that takes the context in which an object appears into account. We do this by combining a PointNet encoder-decoder architecture with a contrastive loss function.
	\item Train our algorithm on the Replica dataset \cite{replica} and analyze the obtained semantic embedding function through a series of experiments.
\end{itemize}


\section{Related Work}
\label{sec:related-work}



Learning vector representations of 3D data has been and still is an active topic of research in the computer vision and machine learning communities. This is driven by the motivation to use representations in various down-stream tasks that require an abstract and generalizable understanding of the entity at hand. Examples for such tasks are classification, scene completion or image inpainting.

A common format for 3D data are voxel grids. This structured representation format lends itself well for applying convolutions. In \cite{sharma16_VConv}, a model is trained to predict voxels that were removed from the original grid. After training, the representations generated by the model can be used for shape recognition or interpolation between instances. Another application for represenations learned from voxel data \cite{girdhar16_TLNet} is the generation of 3D data from 2D images.

Despite these successes, voxel grids have disadvantages compared to point clouds, such as the fixed scale and the need to discretize sensor data before they can be applied. As an alternative, methods directly operating on point cloud data have been investigated, especially after the introduction of PointNet \cite{qi17_PointNet}. Building on top of PointNet, various works attempted to learn meaningful representations in an unsupervised fashion using auto-encoders \cite{chen19_GraphTopology}, generative adversarial networks (GANs) \cite{goodfellow14_gan}, combinations of both \cite{achlioltas18_RepresentationsForPCD,zamorski20_adversarialae} or recurrent neural networks (RNNs) \cite{han18_InterpredictionGAN}. While the representations obtained with these methods were shown to greatly improve down-stream tasks such as reconstruction or classification based on geometry, there is no method that leverages additional information for more semantically meaningful representation. Autoencoders are often used to extract meaningful features of point clouds. To the best of our knowledge, all research in the field analyzes each point cloud separately and ignores context. 

\cite{schmidt2016self} learns dense 3D descriptors for points in 3D spaces by using a contrastive loss. Their goal is to learn a descriptor which is invariant to lighting and viewpoint to be used in finding correspondences between images captured by a depth camera. \cite{florence2018dense} extends this approach to learn more object centric descriptors for use in robotic manipulation.

For this paper, we drew inspiration from a method called word2vec \cite{mikolov13_WordRep} which is a technique from the field of natural language processing. The method is trained to predict a word from its neighbors in a sentence, resulting in vectors that capture semantic information about the original words.

With a similar motivation, in this paper we propose a method to extract semantically meaningful embeddings from 3D point cloud data by not only considering an object's geometric shape, but also its neighborhood in a scene.

\section{Points2Vec}
\label{sec:method}
\begin{figure}
\centering
	\includegraphics[width = 1\linewidth]{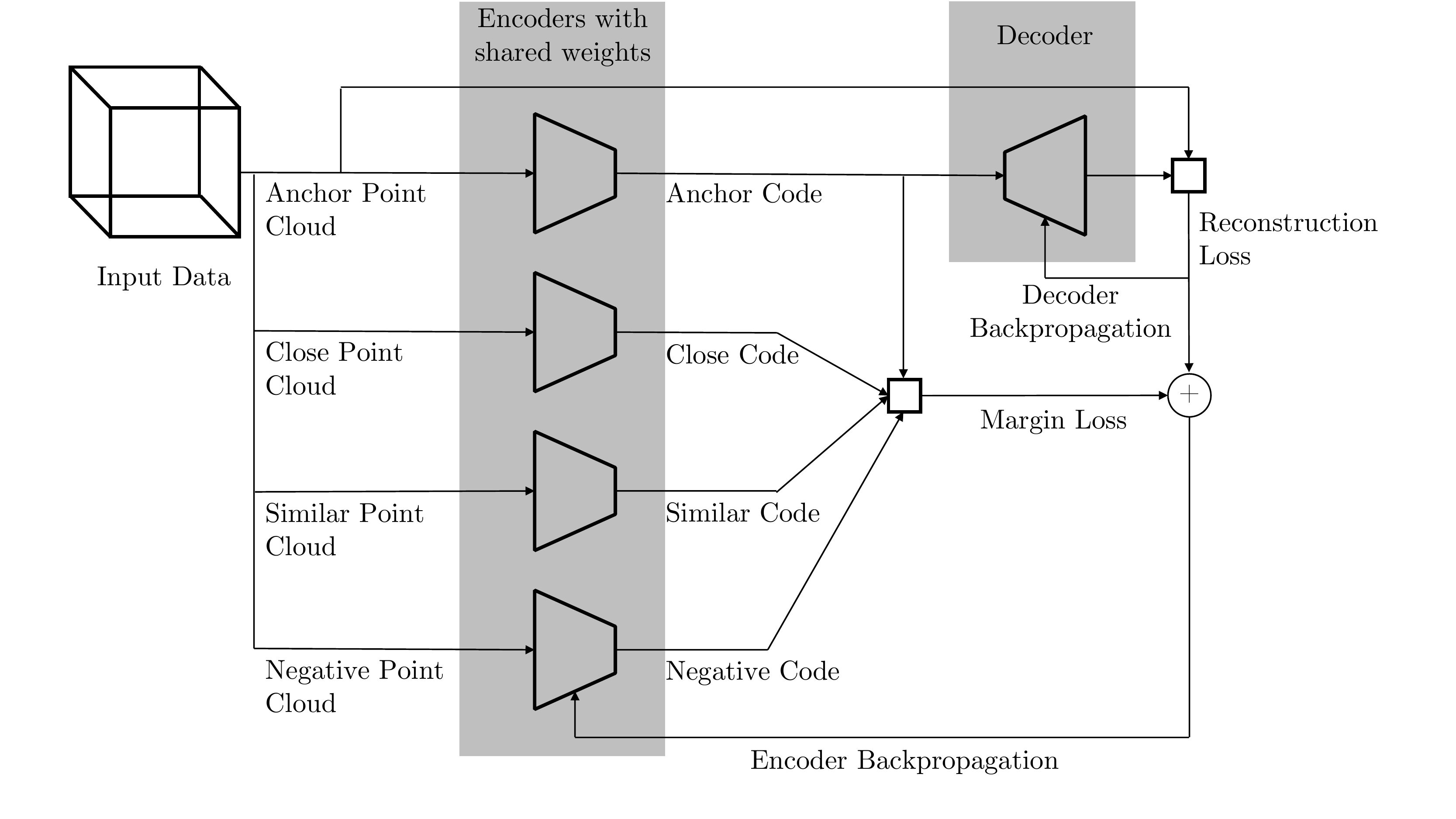}
	\caption{The architecture of our proposed Points2Vec algorithm.}
	\label{fig:margin_p2v}
\end{figure}

In Points2Vec, we extend a PointNet autoencoder with a context module that takes the surrounding objects into account. We conjecture, that objects that are geometrically close and similar in 3D space, should occupy a similar region in latent space. This is enforced by adding a contrastive loss in latent space.
To some extent, this context module is inspired by word2vec \cite{mikolov13_WordRep}, in that it leverages surrounding embeddings to improve the quality of the target embedding.  
Moreover, we assume that adding a loss in latent space improves stability in training. If the only loss function is applied after reconstruction, backpropagation is entirely dependent on the decoder, which may add a significant source of error, particularly with unstructured data. 
For our algorithm, we assume point cloud instances to be segmented, but we do not depend on class labels.
Figure \ref{fig:margin_p2v} shows an overview of our proposed algorithm, which is further described in the following chapters.

\subsection{Autoencoder}
To extract features from the point clouds, we make use of an autoencoder architecture. We use PointNet \cite{qi17_PointNet} as an encoder, which we briefly describe in the following (for the exact network specifications, see Table \ref{tbl:arch} in the Appendix):
The $N\times 3$ point cloud (consisting of $N$ points) gets transformed by means of matrix multiplication with a learnable T-Net. Then, a multi-layer perceptron (MLP) extracts features from each point separately, resulting in an $N\times 64$ matrix, where another transform module is applied. 
Further MLPs convert the matrix to the shape $N\times 512$. Here, we apply the $\max$-operator, reducing the matrix to a vector of size $512$. This vector is then encoded to a lower embedding size using fully connected layers with ReLU activation functions. To avoid overfitting, we employ dropout layers. The embeddings are projected onto the unit hypersphere by normalization. To reconstruct the point cloud from the code, we use a series of fully connected layers with ReLU activation functions.
To account for the unstructured nature of point clouds, the loss function of generative deep neural networks has to be symmetric. To this end, we use the chamfer distance function.

\subsection{Context Module}
We refer to the part of our algorithm that samples context instances and computes the margin loss of the embeddings as context module. 

\begin{figure*}
\centering
	\includegraphics[width = 1\linewidth]{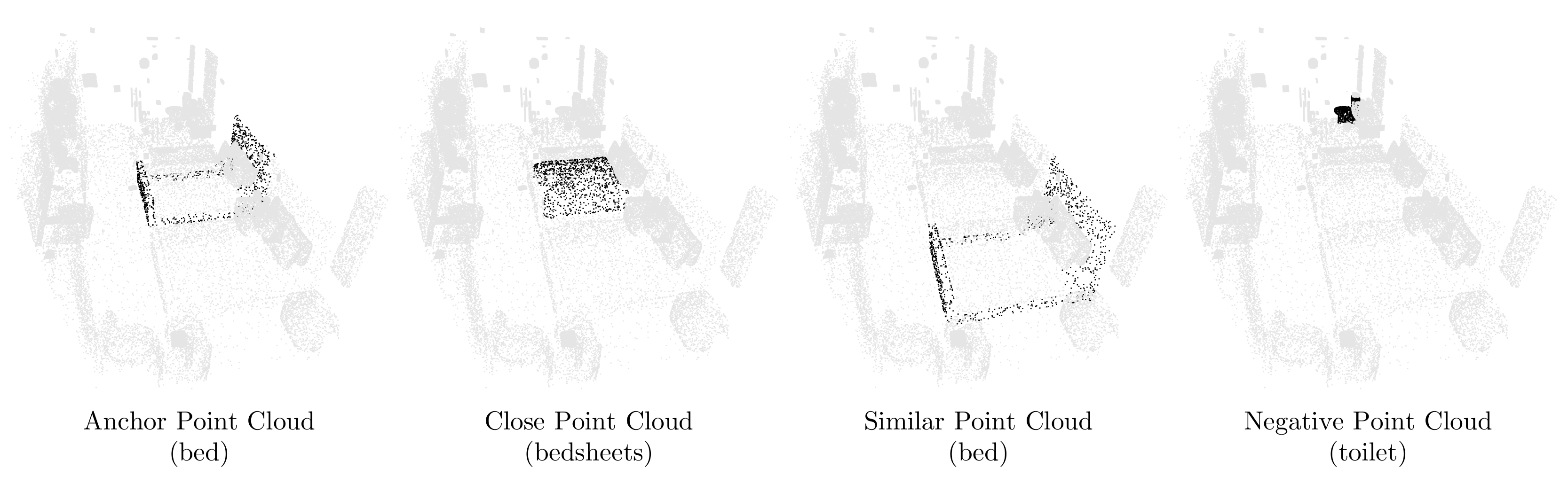}
	\caption{Visualization of the training instance sampling algorithm in Points2Vec. Point clouds in question are marked in black, the rest in grey.}
	\label{fig:sampler}
\end{figure*} 

Our context module is inspired by the triplet margin loss of \cite{schroff15_facenet} and aims at inducing domain knowledge into the hyperspace by computing a contrastive loss of four different embeddings. For this, we make use of the segmentation and run pointclouds through the encoder on a per-instance basis. We sample the following point cloud instances:
\begin{itemize}
	\item \textbf{Anchor Point Cloud:} The anchor point cloud is sampled randomly among all instances. 
	\item \textbf{Close Point Cloud:} We sample an instance situated close to the anchor point cloud in euclidean space. Specifically, we compute the distance to the centroid for all instances and sample a point cloud among the ten closest instances. Sampling is performed by inverse transform sampling, using the distance to the anchor point cloud to inversely weigh the sample probability. The motivation is, as described above, that objects close to each other in euclidean space are often from the same semantic realm (e.g. kitchen utensils or tools in a workshop). Therefore, we aim for the close point cloud embedding to be close to the anchor point cloud embedding.
	\item \textbf{Similar Point Cloud:} Apart from the close instance, we sample an instance similar in euclidean space. To get a similarity measure of point clouds, we use the chamfer distance. To account for different rotations of instances, we pose the task of finding the chamfer distance as an optimization problem. Thus, the goal is to find the angle around the z-axis, for which the chamfer distance is minimal. As with the close point cloud, we sample among the ten most similar instances, using inverse transform sampling to achieve a sample probability inversely proportional to the distance score. Taking similar point clouds into consideration is motivated by the observation that geometrically similar objects are often from the same semantic realm (e.g. they can be used for the same tasks), even if they do not appear close to each other in the observed scene (e.g. cups appearing in the kitchen, the dining room, and the office). Therefore, we aim for the similar point cloud embedding to be close to the anchor point cloud embedding.
	\item \textbf{Negative Point Cloud: } The negative point cloud is sampled at random among all point clouds except the anchor, close and similar point cloud. This is to drive the anchor point cloud embedding away from embeddings of unrelated objects. A random sampling was chosen to reduce the chance of inducing a bias.
\end{itemize}

\begin{figure}
\centering
	\includegraphics[width = 0.9\linewidth]{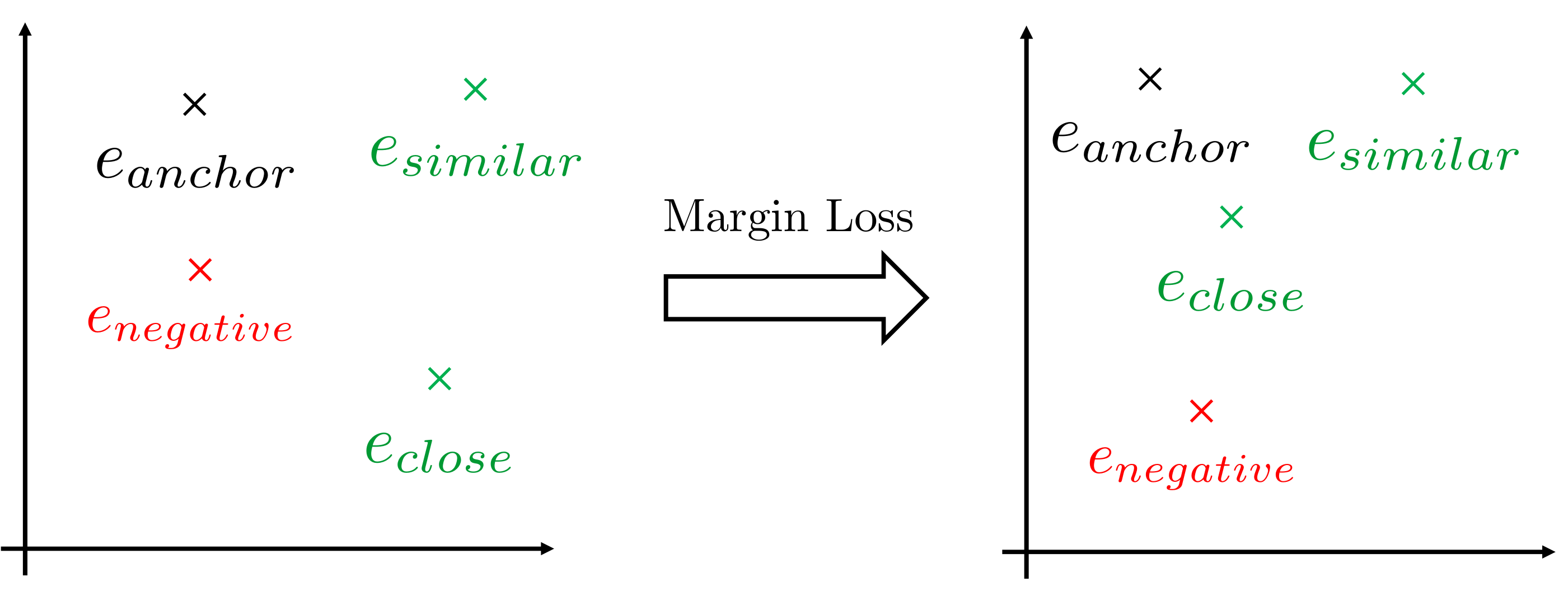}
	\caption{Visualization of the margin loss in latent space. Applying our margin loss pulls $e_\text{similar}$ and $e_\text{close}$ closer to  $e_\text{anchor}$ in latent space pushes $e_\text{negative}$ away from it.}
	\label{fig:margin_loss}
\end{figure}

The sampled point clouds are then run through our encoder, where we apply the following margin loss to the embeddings:
\begin{align}
\mathcal L_{margin} &= \max\left\{\frac{d_{close} + d_{similar}}{2} - d_{negative} + \alpha, 0 \right\} \\
d_{close} &= \left|\left| e_{anchor} - e_{close} \right|\right|^2\\
d_{similar} &= \left|\left| e_{anchor} - e_{similar} \right|\right|^2\\
d_{negative} &= \left|\left| e_{anchor} - e_{negative} \right|\right|^2,
\end{align}
where $\alpha$ is the margin parameter, and $e$ are the embedding vectors. The margin parameter is a tuneable hyperparameter that defines the margin for which a loss is generated. If the two positive embeddings are close, and the negative embedding is far away, the margin function will not produce a loss. In essence, $\mathcal L_{margin}$ achieves to pull close and similar point cloud embeddings closer to the anchor embedding and to push the negative point cloud embedding away from it, as shown in Figure \ref{fig:margin_loss}. For the examples in Figure \ref{fig:sampler}, this would mean that the embedding of the bed sheet and the embedding of the second bed is pulled closer to the embedding of the first bed, whereas the embedding of the toilet is pushed further away from the embedding of the bed. \\

For backpropagation, the margin loss and the reconstruction loss are summed up to update the weights of the encoder. The reconstructor is updated solely using the reconstruction loss. 
At inference time, we only sample the anchor point cloud and run it through the decoder.

\section{Experiments and Results}
\label{sec:experiments}
To test our method, we need a dataset of real world scenes with object segmentation. We use the Facebook Replica dataset \cite{replica}. There would have been larger candidate datasets, but most of them had drawbacks. InteriorNet \cite{InteriorNet18} would have required a dense mapping preprocessing step. SceneNet \cite{SceneNet} is randomly generated and the positions of objects do not make any semantic sense. ScanNet \cite{dai2017scannet} and Matterport3D \cite{Matterport3D} would have been two other datasets we could have used for evaluation.

The Facebook Replica dataset contains 18 photo-realistic indoor scene reconstructions. 
It includes semantic segmentation labels, but we only use this information to evaluate our algorithm. Of the 18 scenes, we use one for validation and one for testing.
The 3D scenes are represented as polygon mesh files. We convert them to point clouds by randomly sampling points on the surface of the mesh. The number of samples on each quadrilateral is proportional to its area. For our experiments, we sampled 1000 points on each instance. To remove uninformative biases, we center all instances in euclidean space.

Several variations to the algorithm were tested, such as a PointNet++ encoder \cite{qi17_pointnetpp} and both deeper and broader encoder and decoder architectures. Moreover, we tested the combination of our autoencoder with an RGB-D inpainting module. To stay concise, we only report the results of our best-performing architecture, as described in Section \ref{sec:method} and in Table \ref{tbl:arch}.

In the following sections, we evaluate the results of the experiments conducted with the proposed algorithm. In essence, there are two outputs that we can evaluate quantitatively, namely the latent-space embeddings and the reconstructions.

\subsection{Latent Space}

\begin{figure}
	\centering
	\includegraphics[width = 0.6\linewidth]{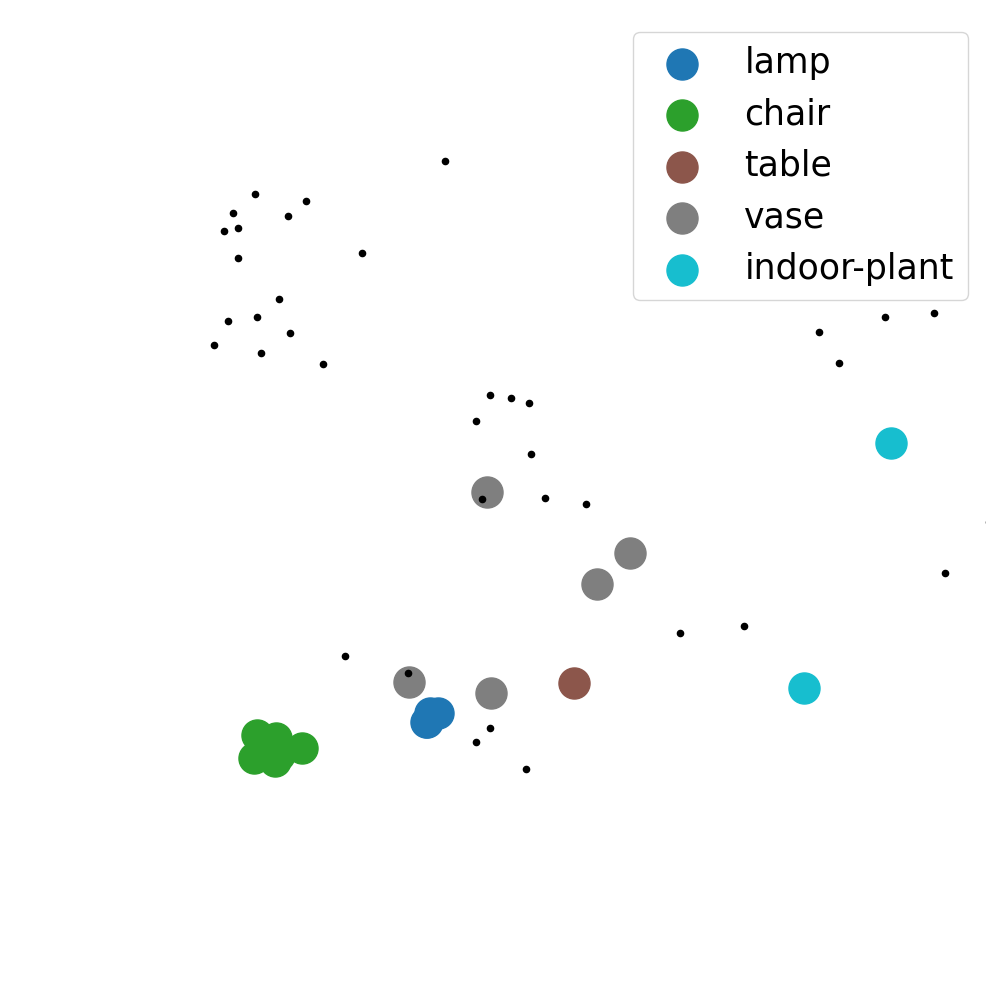}
	\caption{Principal component analysis of the latent space of the test set. A subset of classes is highlighted to show the clustering.}
	\label{fig:latent_space}
\end{figure}

The latent space is not straightforward to evaluate as there is no distinct metric that directly measures the semantic value of the information in the embeddings. Figure \ref{fig:latent_space} shows the latent space of the test set, reduced to two dimensions using principal component analysis. We can observe, that different classes occupy different regions in embedding space. Moreover, classes that we perceive to be semantically similar, such as ``indoor-plant" and ``vase" are close in semantic space.
Two properties that are implied by a rich hyperspace are alignment and uniformity.

\subsubsection{Alignment}
Alignment is achieved when similar features are assigned to similar data inputs, resulting in embeddings of the same classes being close in latent space.
In our algorithm, this is enforced by both losses in the algorithm. With the margin loss, we pull geometrically similar instances closer together and the reconstruction loss implicitly forces geometrically similar instances to occupy a similar region in the embedding space. 
To evaluate alignment, we run two tests over the embeddings: 

First, we run an unsupervised k-means clustering algorithm over the embeddings in hyperspace and evaluate its performance. In an ideal embedding space, each class occupies a particular region in the hyperspace that should be linearly separable from other regions. To quantify the performance of the clustering, we make use of the ground truth labels and compute the adjusted rand index (ARI-KM). This measure takes the permutations of class IDs into account when performing unsupervised clustering. In essence, it can be viewed as a performance measure for a downstream classification task. The ARI-KM scores can be viewed in Table \ref{table:ARI-KM}. We see that our method achieves a higher ARI-KM than a pure autoencoder, both on the test and train set. Moreover, we run a training, where we skip the reconstruction and only backpropagate the margin loss. This results in an ARI-KM index that is lower than in the method that we propose, but outperforms the pure autoencoder. Note that the ARI-KM on the test set is higher than on the train set due to a different number of instances per classes in the two sets.

\begin{table}
\centering
\begin{tabular}{
>{\columncolor[HTML]{C0C0C0}}l |
>{\columncolor[HTML]{FFFFFF}}c |
>{\columncolor[HTML]{FFFFFF}}c |}
\hhline{~|--|}
\multicolumn{1}{c|}{\cellcolor[HTML]{FFFFFF}}                     & \multicolumn{2}{c|}{\cellcolor[HTML]{C0C0C0}ARI-KM}                                                    \\ \hhline{~|--|}
\multicolumn{1}{c|}{\multirow{-2}{*}{\cellcolor[HTML]{FFFFFF}}}   & \multicolumn{1}{l|}{\cellcolor[HTML]{C0C0C0}Train} & \multicolumn{1}{l|}{\cellcolor[HTML]{C0C0C0}Test} \\ \hline
\multicolumn{1}{|l|}{\cellcolor[HTML]{C0C0C0}Points2Vec}          & 0.2331                                             & 0.63                                              \\ \hline
\multicolumn{1}{|l|}{\cellcolor[HTML]{C0C0C0}PointNet Autoencoder} & 0.2246                                             & 0.5178                                            \\ \hline
\multicolumn{1}{|l|}{\cellcolor[HTML]{C0C0C0}Margin Only}         & 0.2278                                             & 0.5491                                            \\ \hline
\end{tabular}
\caption{Evaluation of the unsupervised k-Means clustering in latent space.}
\label{table:ARI-KM}
\end{table}

The second metric for alignment that we extract from two classes, as reported in Table \ref{table:ACD}. For this, we take all possible combinations of two instances from two classes, compute each cosine distance and average it over the combinations. We perform these calculations both for two different classes, as well as for all combinations of instances of the same class. In a well-clustered latent space, we expect the averaged cosine distance between instances of the same class to be lower than between two different classes.
Not only can we see that intra-class cosine distances are lower than inter-class distances, but we also get a sense of the distances between two instances of different classes. For example, the averaged distance between all windows and all blinds is relatively small, indicating closeness in semantic space. This is consistent with our semantic understanding of the two objects.
Moreover, if we average all intra-class distances, we get a lower distance (0.262) than if we average all inter-class distances (0.997).

\begin{table}
\centering
\begin{tabular}{l|c|c|c|c|}
\hhline{~|----|}
\multicolumn{1}{c|}{\cellcolor[HTML]{FFFFFF}}& \multicolumn{4}{c|}{\cellcolor[HTML]{C0C0C0}Averaged cosine distances} \\  
\hhline{~|----|}
\multicolumn{1}{c|}{} & \cellcolor[HTML]{C0C0C0}Chair & \cellcolor[HTML]{C0C0C0}Plate & \cellcolor[HTML]{C0C0C0}Blinds & \cellcolor[HTML]{C0C0C0}Window \\ \hline
\multicolumn{1}{|l|}{\cellcolor[HTML]{C0C0C0}{\color[HTML]{000000} Chair}}  & 0.001 & 0.924 & 0.863 & 1.13  \\ \hline
\multicolumn{1}{|l|}{\cellcolor[HTML]{C0C0C0}{\color[HTML]{000000} Plate}}  & 0.924 & 0.186 & 1.255 & 1.289 \\ \hline
\multicolumn{1}{|l|}{\cellcolor[HTML]{C0C0C0}{\color[HTML]{000000} Blinds}} & 0.863 & 1.255 & 0.113 & 0.137 \\ \hline
\multicolumn{1}{|l|}{\cellcolor[HTML]{C0C0C0}{\color[HTML]{000000} Window}} & 1.13  & 1.289 & 0.137 & 0.035 \\ \hline
\end{tabular}
\caption{Averaged cosine distances between all instances of two classes on the test set. Intra-class distances are lower, indicating alignment in latent space.}
\label{table:avg_cos_distances}
\end{table}

\subsubsection{Uniformity}
\begin{figure}
	\centering
	\includegraphics[width = 0.7\linewidth]{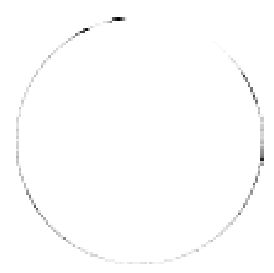}
	\caption{Heatmap of the hyperspace after principal component analysis and normalization. Uniform distribution of feature vectors indicates that maximal information is preserved.}
	\label{fig:heatmap}
\end{figure}

Second, we evaluate uniformity. Uniformity describes the trait that features are evenly distributed in latent space. Uniform distribution preserves maximal information in the latent space, as shown in \cite{wang20_ContrastiveRepLearning}.
Figure \ref{fig:heatmap} shows a histogram of the embeddings. For this visualization, we reduced the dimensionality of the hyperspace from 256 dimensions down to two using principal component analysis and normalized the embeddings. As is visible, the embeddings are evenly distributed along the circle. We conjecture that the areas of higher density are due to class imbalance.

\begin{figure}
	\centering
	\includegraphics[width = 1\linewidth]{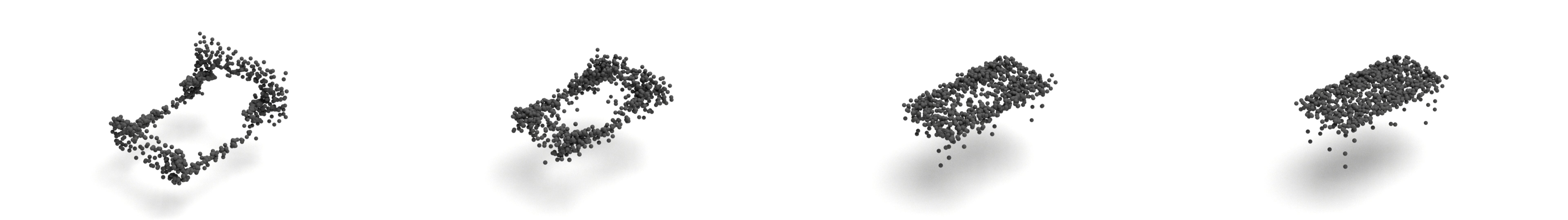}\\
	\includegraphics[width = 1\linewidth]{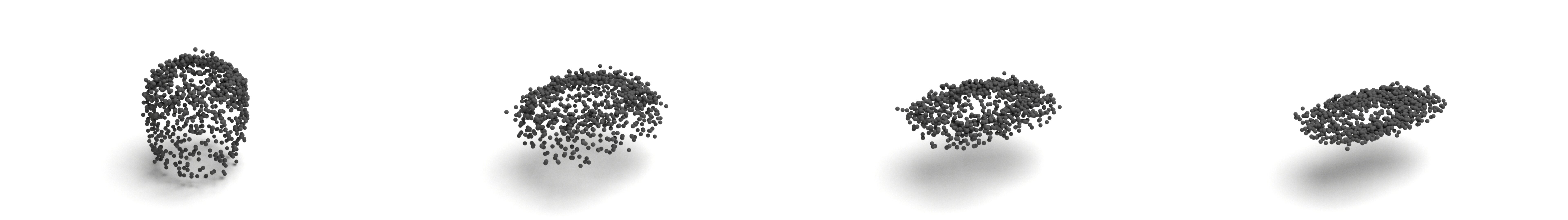}\\
	\includegraphics[width = 1\linewidth]{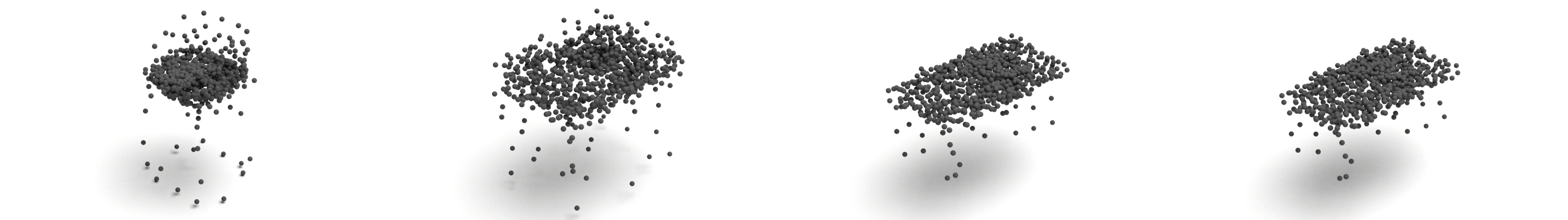}\\
	\caption{Interpolation between two different classes.}
	\label{fig:interpolations_bed_table}
\end{figure}

Further, we test uniformity by interpolating and reconstructing embeddings. A uniform distribution in latent space should result in a smooth transition of the reconstructed point clouds. Results of this are shown in Figure \ref{fig:interpolations_bed_table}. Here, we interpolate between two instances of different classes, namely a bed and a table. We observe a smooth transition between the two classes.

\subsubsection{Robotics Use-Case}
To give an example of a use-case in robotics, we give our algorithm the task of finding semantically similar objects in an unseen environment. In the example shown in Figure \ref{fig:use_case}, we search for objects that are semantically similar to a chair. With our method, the most semantically similar instances in the scene are all the other chairs and the table. This showcases the principle of our method. The chairs are perceived as semantically similar, as they have a similar shape. The table, on the other hand, doesn't have a similar shape but regularly co-occurs along with tables in our training set. The margin loss thus pulls its embedding closer to the embedding region of the chairs. When running the same experiment with a pure PointNet autoencoder, the table is not among the most similar instances, even though we would intuitively associate a table with chairs.

\subsubsection{Reconstruction}
\begin{figure}
	\centering
	\includegraphics[width = 0.9\linewidth]{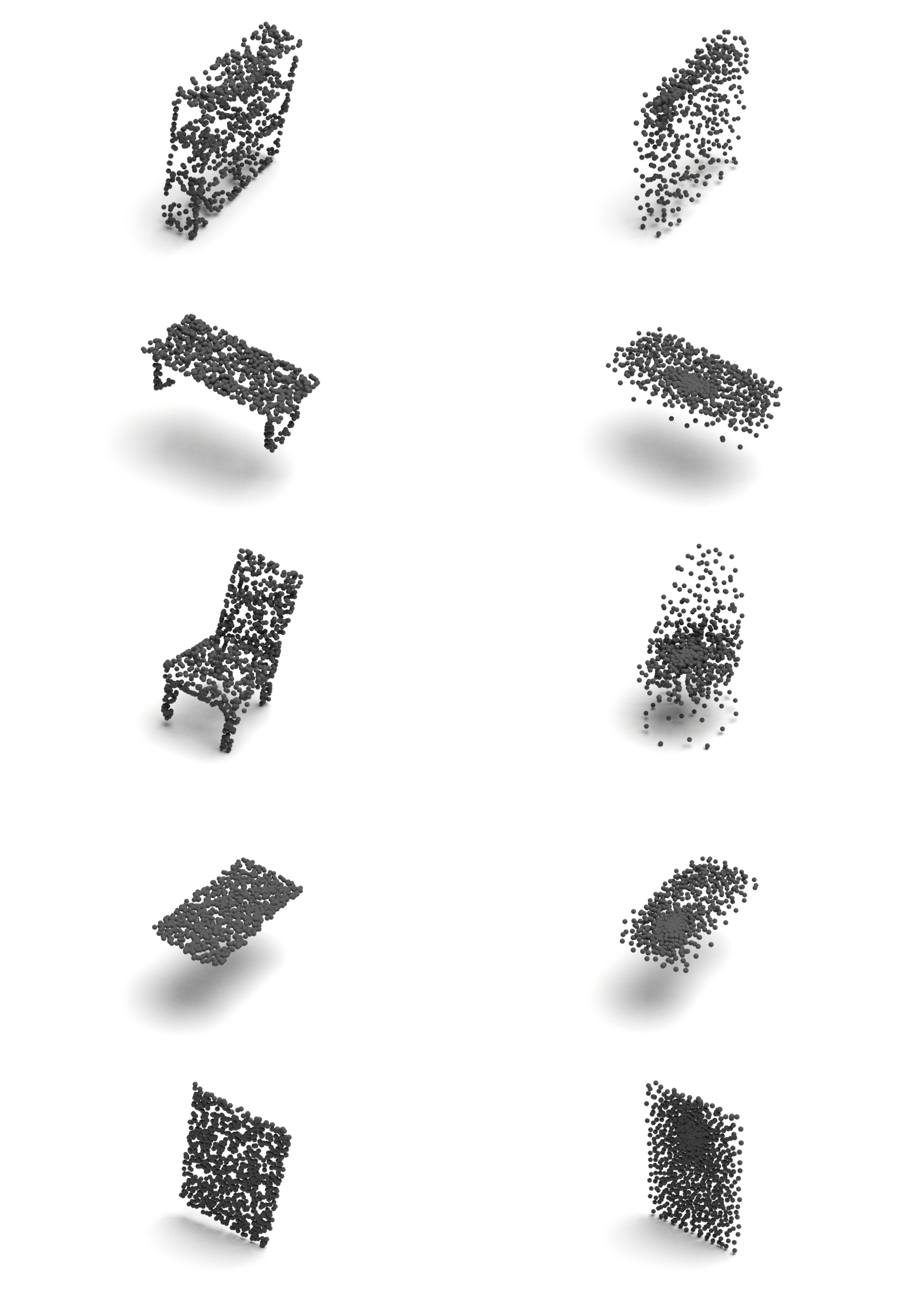}
	\caption{Ground truth (left) and reconstructions (right) of unseen instances}
	\label{fig:reconstruction_pairs.png}
\end{figure}

In this section, we evaluate the reconstructions of our algorithm. While achieving high-quality reconstructions is not the actual goal of this method, it is still a viable quantitative measure for the semantic richness of the feature vectors. Figure \ref{fig:reconstruction_pairs.png} shows combinations of ground truth - reconstruction pairs from the unseen test set.

To measure the quality of the reconstruction, we compute the chamfer distance between the ground truth and the reconstruction. 
On Table \ref{table:ACD}, we list the resulting averaged chamfer distance (ACD) measures on the train and test set. We compare our approach to the PointNet autoencoder in order to evaluate how adding our context module affects reconstruction. 
We see that both on the training and on the test set, our method achieves a lower ACD, indicating a better reconstruction. On the test set, we observe an improvement of 8\% compared to the PointNet autoencoder.

\begin{table}
\centering
\begin{tabular}{l|c|c|}
\hhline{~|--|}
\multicolumn{1}{c|}{} & \multicolumn{2}{c|}{\cellcolor[HTML]{C0C0C0}ACD} \\ 
\hhline{~|--|}
 & \multicolumn{1}{l|}{\cellcolor[HTML]{C0C0C0}{\color[HTML]{000000} Train}} & \multicolumn{1}{l|}{\cellcolor[HTML]{C0C0C0}{\color[HTML]{000000} Test}} \\ \hline
\multicolumn{1}{|l|}{\cellcolor[HTML]{C0C0C0}{\color[HTML]{000000} Points2Vec}} & 0.0654 & 0.0946 \\ \hline
\multicolumn{1}{|l|}{\cellcolor[HTML]{C0C0C0}{\color[HTML]{000000} PointNet Autoencoder}} & 0.067 & 0.103 \\ \hline
\end{tabular}
\caption{Averaged chamfer distances (ACD) on training and test set for Points2Vec and PointNet autoencoder. Our method achieves a lower ACD than a PointNet autoencoder, indicating better reconstruction.}
\label{table:ACD}
\end{table}

\section{Conclusions}
\label{sec:conclusions}

We presented an algorithm for unsupervised learning of object representations from point clouds. To the best of our knowledge, this is the first algorithm that approaches the task of extracting semantic feature vectors of point clouds by not just looking at the isolated point cloud, but by taking the context in which it appears into account. 

We learned and evaluated our algorithm on the Replica dataset. For evaluation, we tested our method on unseen data and evaluated the latent space for alignment by running an unsupervised k-Means clustering algorithm over the embeddings and quantifying the performance using ground truth annotation. Moreover, we were able to reconstruct realistic point cloud instances. For benchmarking, we run a PointNet autoencoder, as proposed in \cite{achlioltas18_RepresentationsForPCD} on the same dataset.  Our method achieves a better score both in latent space clustering and in reconstrucion than our benchmark, the regular PointNet autoencoder. These results encourage our hypothesis, that context can add semantic information in point cloud feature learning. Due to better clustering in latent space, less data annotation is required in a downstream classification task.

For now, our method runs on instance-segmented point clouds. In a real-life application, it might therefore be combined with a point cloud object segmentation algorithm, that segments object instances from pointclouds, such as \cite{ding2019votenet} or \cite{Pham2017}.

Our algorithm is a first proposal of what an algorithm that learns semantic representations from real-world scenes can achieve. Future work might make use of generative adversarial networks \cite{goodfellow14_gan} to make better use of the data and better encode the geometry of the objects. A NeuralSampler \cite{remelli19_neuralsampler} could be used to deal with the fact that not all objects are equally well represented with the same number of points. 

We ran our algorithm on a small scale dataset consisting of 18 indoor spaces with limited range. Our work is inspired by methods in natural language processing which are learned using datasets with many orders of magnitude more data. 3D sensors are rapidly becoming more common in human environments, as exemplified by the release of the iPhone 12 Pro with a Lidar sensor. Our promising early results are an encouraging sign that context provides valuable semantic information in unsupervised point cloud feature learning. We have little doubt that the quality and range of the learned representations would only improve when learning from larger, more varied, datasets. Imagine what we could learn, if our dataset contained 1 million scanned scenes.

\section*{Appendix}
\subsection*{Additional implementation details}

\begin{table}[h]
\centering
\begin{tabular}{l|c}
    Learning Rate & 0.0001 \\
    Code Size & 256 \\
    Batch Size & 20 \\
    Number of epochs &  4000 \\
    Margin Loss Parameter & 1 \\
    Margin Loss to Reconstruction Loss Ratio & 10
\end{tabular}
\caption{Hyperparameters used in the experiments.}\label{tbl:arch}
\end{table}

\begin{table}[h]
\centering
\begin{tabular}{l|c|l}
\cline{2-2}
 & \cellcolor[HTML]{C0C0C0}\textbf{Layers} &  \\ \cline{2-2}
 & 3x3 Transform &    \\ \cline{2-2}
 & \makecell{3x64 MLPs with shared weights,\\Batch Normalization} &   \\ \cline{2-2}
 & 64x64 Transform &   \\ \cline{2-2}
 & \makecell{3x64 MLPs with shared weights,\\Batch Normalization} &   \\ \cline{2-2}
 & \makecell{64x128 MLPs with shared weights,\\Batch Normalization} &  \\ \cline{2-2}
 & \makecell{128x1024 MLPs with shared weights,\\Batch Normalization} &  \\ \cline{2-2}
 & max-Pool &  \\ \cline{2-2}
 & 1024x512 MLP &  \\ \cline{2-2}
 & 512x256 MLP & \\ \cline{2-2}
 & 256x256 MLP &   \\ \cline{2-2}
 & Code &    \\ \cline{2-2}
 & 256x256 MLP &   \\ \cline{2-2}
 & 256x512 MLP &   \\ \cline{2-2}
 & 512x1024 MLP &   \\ \cline{2-2}
 & 1024x3072 MLP &    \\ \cline{2-2}
\end{tabular}
\caption{Encoder/Decoder network architecture.}\label{tbl:arch}
\end{table}

\bibliography{Paper/references}
\bibliographystyle{ieeetr}

\end{document}